# Unified CNNs and transformers underlying learning mechanism reveals multi-head attention modus vivendi

Ella Koresh[a,1], Ronit D. Gross[a,1], Yuval Meir[a,1], Yarden Tzach[a], Tal Halevi[a], and Ido Kanter[a,b,*]

[a]Department of Physics, Bar-Ilan University, Ramat-Gan, 52900, Israel.
[b] Gonda Interdisciplinary Brain Research Center, Bar-Ilan University, Ramat-Gan, 52900, Israel.

*Corresponding author at: Department of Physics, Bar-Ilan University, Ramat-Gan, 52900, Israel. E-mail address: ido.kanter@biu.ac.il (I. Kanter).

[1]These authors equally contributed to this work

## Abstract

Convolutional neural networks (CNNs) evaluate short-range correlations in input images which progress along the layers, whereas vision transformer (ViT) architectures evaluate long-range correlations, using repeated transformer encoders composed of fully connected layers. Both are designed to solve complex classification tasks but from different perspectives. This study demonstrates that CNNs and ViT architectures stem from a unified underlying learning mechanism, which quantitatively measures the single-nodal performance (SNP) of each node in feedforward (FF) and multi-head attention (MHA) sub-blocks. Each node identifies small clusters of possible output labels, with additional noise represented as labels outside these clusters. These features are progressively sharpened along the transformer encoders, enhancing the signal-to-noise ratio. This unified underlying learning mechanism leads to two main findings. First, it enables an efficient applied nodal diagonal connection (ANDC) pruning technique without affecting the accuracy. Second, based on the SNP, spontaneous symmetry breaking occurs among the MHA heads, such that each head focuses its attention on a subset of labels through cooperation among its SNPs. Consequently, each head becomes an expert in recognizing its designated labels, representing a quantitative MHA modus vivendi mechanism. This statistical mechanics inspired viewpoint enables to reveal macroscopic behavior of the entire network from the microscopic performance of each node. These results are based on a compact convolutional transformer architecture trained on the CIFAR-100 and Flowers-102 datasets and call for their extension to other architectures and applications, such as natural language processing.

1. **Introduction**

Three years after the introduction of transformer encoders[1], the vision transformer (ViT)[2] was proposed in 2020 as an alternative to convolutional neural networks (CNNs)[3-5] for image classification tasks. ViT architectures typically employ repeated transformer encoder blocks, each comprising multi-head attention (MHA) and feedforward (FF) sub-blocks. While each transformer encoder captures long-range correlations across the entire input image, convolutional layers (CLs) progressively expand their receptive field[6], starting from local correlations and gradually incorporating broader contextual information. Simulation results suggests that state-of-the-art ViT and CNN architectures achieve comparable accuracies on small and medium-sized datasets. However, the belief is that ViT models typically outperform CNNs on large datasets[7, 8]. This superior performance is often attributed to ViT's ability to learn long-range correlations through its transformer blocks and its utilization of numerous learnable parameters which can exceed billions. This has led to the prevailing view that the learning mechanisms underlying ViT and CNNs are fundamentally distinct.

Contrary to this prevailing view, we demonstrate that the underlying learning mechanism of a compact ViT architecture shares striking similarities with that of CNN architectures. The presented analysis centers on a novel method for quantifying single-nodal performance (SNP) within each transformer encoder, encompassing both MHA and FF sub-blocks (Fig. 1). By examining SNP, the underlying learning mechanism of MHA is uncovered, characterized by spontaneous symmetry breaking among the attention heads. Specifically, each head specializes in classifying a subset of labels, a phenomenon that becomes increasingly pronounced in the final encoder blocks. Moreover, this understanding of the SNP mechanism enables the development of a quantitative pruning technique that significantly reduces the model size of a compact ViT architecture without compromising accuracy. Findings are primarily presented using a compact convolutional transformer, CCT-7/3x1[9], comprising seven transformer encoders (Fig. 1) and a single CL as a tokenizer. This model achieves an accuracy of 0.809 on the CIFAR100[10] dataset after 300 training epochs. Supporting evidence is also provided from the CCT-7/7x2 architecture[9] trained on the Flowers-102 dataset[11], where it attains an accuracy of ~0.972 after 300 training epochs.

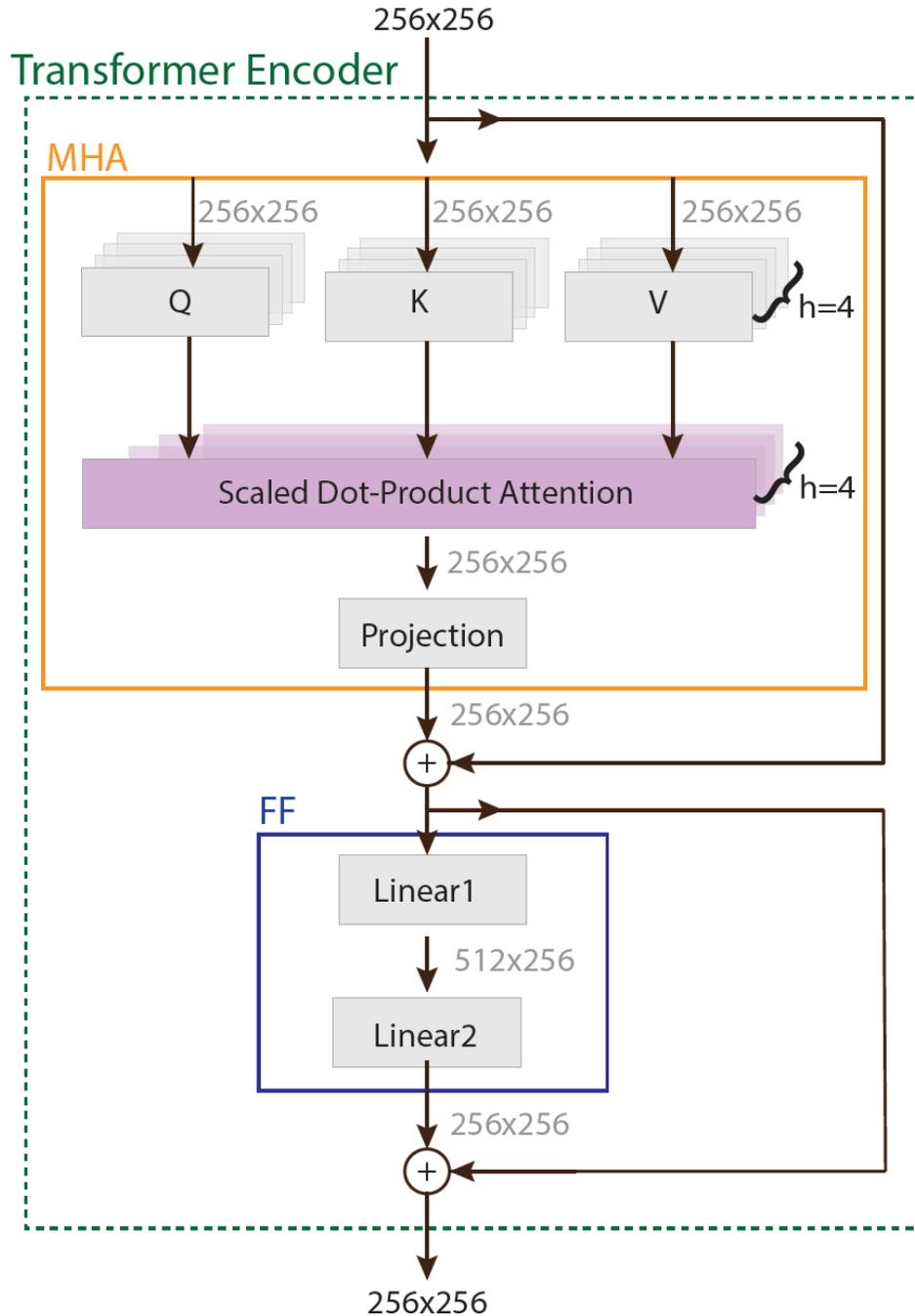

**Fig. 1.** Scheme of a transformer encoder of the CCT-7/3x1 architecture, consisting of two sub-blocks, MHA and FF, where FC layers are denoted by gray rectangles.

## 2. Underlying ViT learning Mechanism

The underlying successful learning mechanism of ViT hinges on quantifying the SNP for each node within the network. This analysis is demonstrated using the CCT-7/3x1[9] architecture, which employs a CL as a tokenizer instead of image patching, followed by seven transformer encoders and a classifier head composed of sequence pooling and a fully connected (FC) layer to the output layer (Fig. 2). To analyze SNP, a CCT-7/3x1 architecture pre-trained on the CIFAR-100 dataset is utilized. Initially, the weights of the first $m$ trained transformer encoder blocks remain unchanged (frozen). Their output units, $256 \times 256$, represent the preprocessed input image after passing through this partial CCT-7/3x1 architecture. These units are then connected to a randomly initialized classifier head, comprising sequence pooling and an FC layer with 100 outputs. This classifier head is trained on CIFAR-100, to minimize the loss, a relatively simple computational task (Fig. 2, middle). The accuracy of this partial CCT-7/3x1 architecture, with only $m$ transformer encoders and its newly trained classifier head, progressively increases with $m$ (Table 1). Notably, for $m = 7$, the accuracy slightly surpasses that of the original pre-trained model, increasing from $0.809$ to $0.811$. This trained classifier head (Fig. 2, middle) is then used to quantify the functionality of each the $256$ output units of the $m^{th}$ transformer encoder layer, defining the SNP.

To calculate SNP, all weights in the FC layer of the trained classifier head are silenced except for those (100) connected to a specific node (Fig. 2, right). Next, the image validation set is then presented to the network and processed by the first $m$ transformer encoders and sequence pooling layer. The output units are influenced solely through the limited aperture of the single node, summarized by a $100 \times 100$ matrix (Fig. 2, right and Fig. 3a). Each element $(i, j)$ of this matrix represents the average field generated on output unit $j$ by validation inputs with label $i$, normalized by the maximum element in the matrix (Fig. 3a, left). A Boolean clipped matrix is then derived by a given threshold (Fig. 3a, middle), followed by permutation to form diagonal clusters (Fig. 3a, right). Above-threshold elements outside these diagonal clusters are classified as noise, $n$ (yellow elements in Fig. 3a, right). The average properties of the $256$ SNP matrices, each of size $100 \times 100$, are summarized in Table 1.

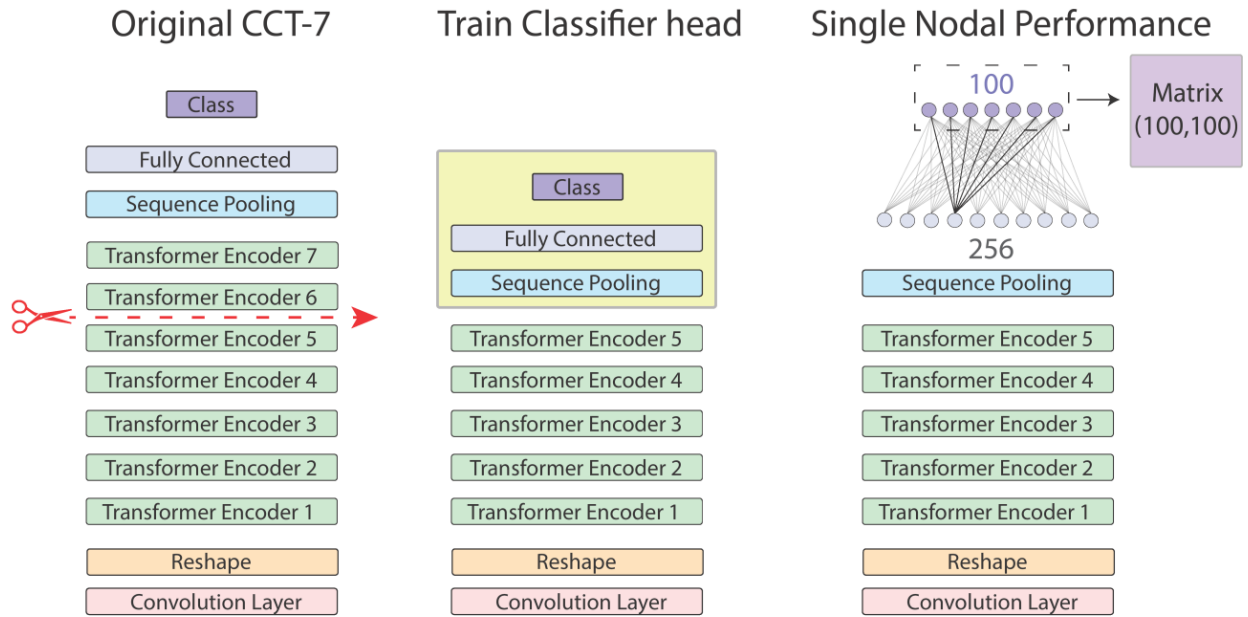

**Fig. 2.** Methodology for measuring single nodal performance (SNP), exemplified on the CCT-7/3x1 architecture. **Left:** A pre-trained CCT-7/3x1 on CIFAR-100 is cut after the $m^{th}$ transformer encoder (exemplified by $m = 5$). **Middle:** A classifier head (yellow) is trained on CIFAR-100 to minimize the loss function. **Right:** All weights of the FC layer are silenced (light gray) except for the emerging 100 weights from the selected node (black). The 100 output fields, averaged over each validation inputs' label, generate a $100 \times 100$ matrix which is used to quantify this SNP.

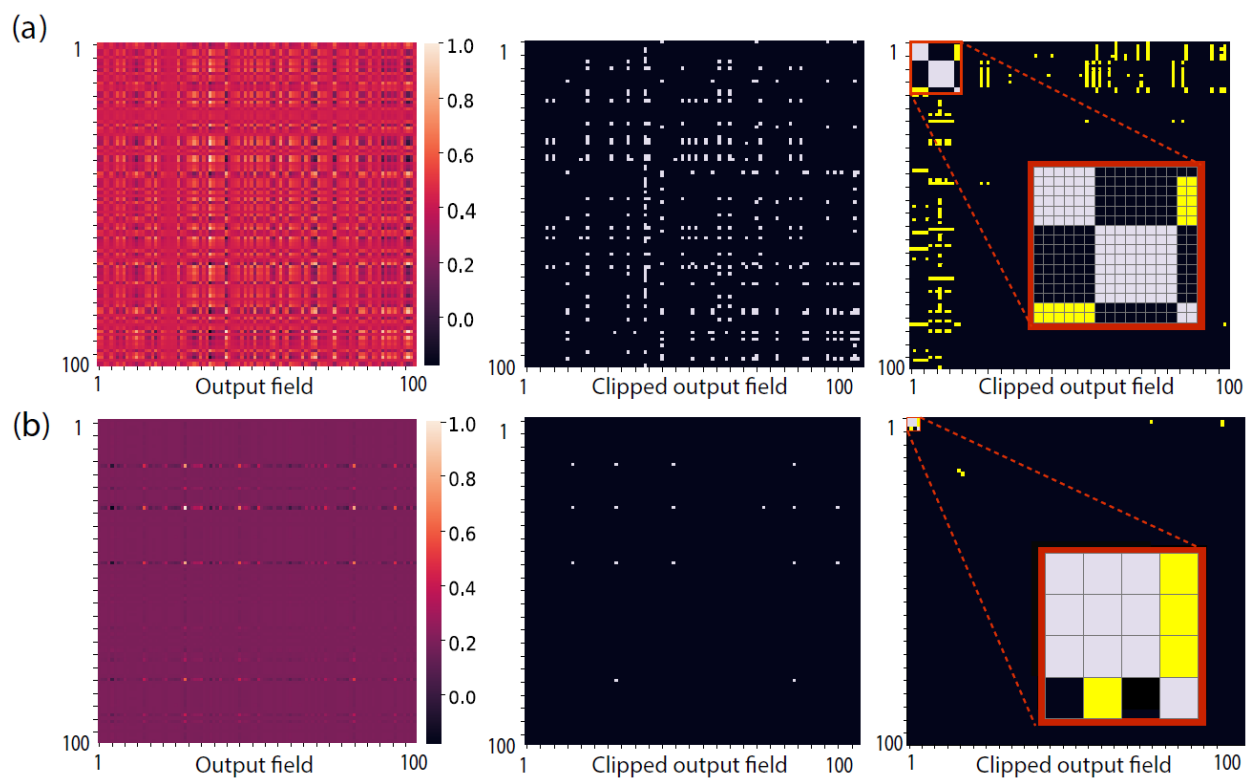

**Fig. 3.** Single Nodal performance (SNP). **(a) Left:** A nodal matrix generated by one of the 256 input units to the FC layer of the classifier head of the 7[th] transformer encoder, connected to 100 output units, where all other connections are silenced. The element ($i$, $j$) stands for the averaged fields generated by label $i$ validation inputs on an output unit $j$, where the matrix elements are normalized by their maximal element. **Middle:** The Boolean clipped matrix following threshold 0.3. **Right:** Permutations of the clipped matrix resulting in block diagonal clusters of sizes $6, 8$ and $2$, and above-threshold elements out of the clusters denoted by yellow. **(b)** Similar to panel a, where a classifier head is trained on the output of Linear1 of the 7[th] transformer encoder, consisting of 512 hidden units.

| CCT-7/3x1 on CIFAR-100 | | | | | |
|---|---|---|---|---|---|
| Transformer | $Acc$ | $N_c$ | $C_s$ | $diag$ | $n$ |
| 7 | 0.811 | 4.9 | 2.9 | 14.6 | 417.2 |
| 6 | 0.797 | 3.6 | 1.8 | 6.8 | 299.5 |
| 5 | 0.767 | 3.3 | 2.0 | 6.9 | 318.5 |
| 4 | 0.729 | 3.5 | 2.2 | 7.8 | 346.9 |
| 3 | 0.667 | 3.5 | 2.5 | 9.1 | 411.9 |
| 2 | 0.576 | 3.4 | 2.8 | 9.5 | 475.4 |
| 1 | 0.464 | 3.3 | 2.8 | 9.3 | 517.5 |

**Table 1.** CCT-7/3x1 trained on CIFAR-100 where accuracy per transformer encoder, $ACC$, is measured using a trained classifier head connected to the output of Linear2 of each transformer block (Fig. 2, right). The statistical features of the 256 SNP matrices of the input units to the FC layer of each one of these trained classifier heads are summarized by the average number of clusters per matrix, $N_c$, average cluster size, $C_s$, average diagonal, $diag \sim N_c \cdot C_s$, and average number of noise elements per matrix, $n$ (numbers are rounded).

A similar procedure is applied to the FF sub-block of the seventh transformer encoder, which comprises two FC layers[1], denoted as Linear1 and Linear2 (Fig. 1). For example, for the output of Linear1 (input to Linear2), comprising 512 hidden units, the sequence pooling of the classifier head results in 512 hidden units fully connected to the 100 output units. The SNP of each of these 512 units is summarized in a $100 \times 100$ matrix, typically exhibiting smaller clusters and a lower noise level (Fig. 3b), and their statistics are $C_s = 1.65$ for the average cluster size, $N_c = 1.34$ for the average number of clusters, $diag \sim N_c \cdot C_s = 2.21$, and $n = 58$ for the average number of noise elements per matrix. Note that the diagonal is explicitly calculated for all matrices and then averaged, which approximately equals to $N_c \cdot C_s$.

The significance of these SNP clusters (Fig. 3) lies in their information-theoretic implications. The emergence of small clusters indicates that each node effectively identifies a small subset of possible output labels among the 100 labels. An input with a label belonging to an SNP cluster generates strong output fields on the corresponding output units belonging to the cluster, whereas an input with a label outside the cluster produces weak fields across all outputs (neglecting noise). Thus, a single node reduces the predicted input labels from 100 to the size of its cluster, effectively filtering out irrelevant labels, where the noise elements (Fig. 3, right matrices) are negligible. To understand the underlying learning mechanism of ViT, the interplay between the number and size of clusters and the fraction of noise elements in the SNP matrices is quantified.

The statistics of the 256 SNP matrices (Fig. 3a) reveal that each label appears almost equally within the clusters (as illustrated in Fig. 4a). Furthermore, the fields induced by the validation set on the 100 output units within the diagonal clusters' elements are also nearly uniform (Fig. 4b). These observations suggest that the learning process ensures balanced influence from the SNP clusters on all 100 outputs, preventing any label dominance. These statistical features (Fig. 4), along with the average cluster size ($C_s$), average number of clusters per matrix ($N_c$), and average number of noise elements per matrix ($n$) (Table 1) allow for a quantitative analysis of the underlying transformer encoder learning mechanism, analogous to single filter performance in deep CNNs[12, 13].

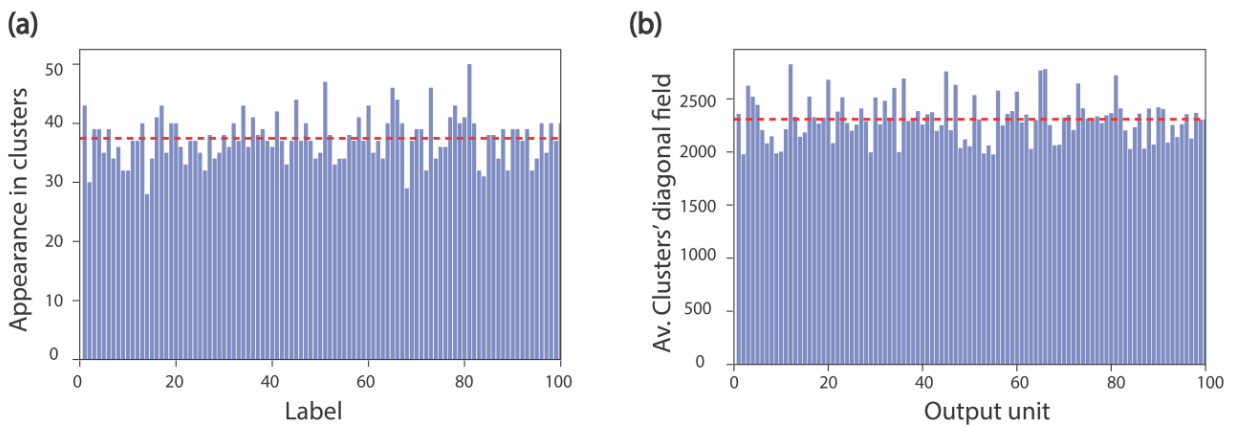

**Fig. 4.** Statistical features of the 256 SNP matrices corresponding to the input units to the FC layer of the classifier head of the 7[th] transformer encoder of CCT-7/3x1. **(a)** Number

of appearances of each label ($Signal$) in the diagonal clusters of the SNP matrices and their average value (dashed-red horizontal line). **(b)** Total field on each output unit, representing a label, generated by the diagonal cluster elements, and their average value (dashed-red horizontal line).

The average number of appearances of each label (representing the signal) within in the clusters of the $N_M = 256$ SNP matrices of the seventh transformer encoder output is

$$signal = (C_s \cdot N_c \cdot N_M)/N_l \quad (1)$$

resulting in 37.4 average signal per label (Table 1). The average internal cluster noise ($noise_I$) is calculated as the average number of appearances of other labels within the clusters forming the $signal$ of a given label,

$$noise_I = \frac{(C_s - 1)}{N_l - 1} \cdot signal \quad (2)$$

which results in $noise_I \sim 0.73$, where $SNR_I = \frac{signal}{noise_I} \gg 1$.

The second type of noise arises from the average number of above-threshold matrix elements outside the clusters, $n$ (Fig. 3, right matrices). Assuming uniform noise distribution across off-diagonal matrix elements, the average noise per matrix element is

$$noise_E = n \cdot \frac{N_M}{(N_l)^2}, \quad (3)$$

where the denominator $(N_l)^2 - (C_s)^2 \cdot N_c$ is approximated by $(N_l)^2 = 10{,}000$ (9,958 in the example of Table 1). The total signal-to-noise-ratio (SNR) is then calculates as

$$SNR = \frac{signal}{noise_I + noise_E} = \frac{C_s \cdot N_c \cdot N_M \cdot N_l}{N_l \cdot (C_s - 1) \cdot signal + n \cdot N_M} \quad (4)$$

where $N_l - 1 \sim N_l$. The SNR increases along the transformer encoder blocks (Table 2) and constitutes the foundation of the mechanism underlying successful ViT learning. Because $noise_I \ll noise_E$ (Table 2), the denominator of Eq. (4) is approximately proportional to the noise, $n$, generally decreases along the transformer encoder blocks (Table 1),

$$SNR \approx \frac{C_s \cdot N_c \cdot N_l}{n} \quad (5)$$

which contributes to the increase in SNR. The increase in $n$ in the seventh transformer encoder (Table 1) is compensated by the increased average diagonal, leading to an enhanced SNR.

The reported results (Fig. 4 and Tables 1 and 2) are robust to the selected threshold within the range $[0.2, 0.4]$. However, for significantly higher thresholds, the monotonicity of the SNR is violated (Table 2). Because the values of the diagonal elements are typically the highest in the matrix (Fig. 3), as the threshold increases, $noise_E$ (Eq. 3) decreases significantly faster than $diag$ (Table 1), constituting the $signal$ (Table 1 and Fig. 4a), where $C_s$ converges toward unity. With such high threshold limits, the SNR, calculated using averaged statistics of the $256$ matrices (Table 1 and 2), diverges (Figs. 5a and 5c), but its interpretation becomes questionable owing to the loss of uniformity among labels (Fig. 4a). Therefore, SNR is replaced by the minimal SNR, $SNR_{min}$, based on the label with the fewest appearances in matrices' diagonals. The local non-monotonicity of $SNR_{min}$ as a function of threshold (Fig. 5b and 5d) is a result of its fluctuations in a finite matrix sizes. For the seventh transformer encoder $SNR_{min}$ as a function of the threshold reveals that the maximum $SNR_{min}$ occurs at a threshold of ~ 0.3, and for the sixth encoder at ~ 0.15 (Figs. 5b and 5d). Consequently, excessively high thresholds ($SNR_{min} < 1$) are excluded. However, the possibility of optimizing the SNR using varying relatively small thresholds per encoder block warrants further investigation.

| CCT-7/3x1 Transformer SNR | | | | |
|---|---|---|---|---|
| Transformer | $signal$ | $noise_I$ | $noise_E$ | $SNR$ |
| 7 | 37.4 | 0.73 | 10.6 | 3.28 |
| 6 | 17.6 | 0.15 | 7.6 | 2.25 |
| 5 | 17.7 | 0.18 | 8.1 | 2.12 |
| 4 | 20.1 | 0.24 | 8.8 | 2.20 |
| 3 | 23.3 | 0.36 | 10.5 | 2.13 |
| 2 | 24.5 | 0.44 | 12.1 | 1.94 |
| 1 | 23.8 | 0.43 | 13.2 | 1.74 |

**Table 2.** CCT-7/3x1 trained on CIFAR-100 (Table 1) and the resulted $signal$ (Eq. (1)), $noise_I$ (Eq. (2)), $noise_E$ (Eq. (3)) and SNR (Eq. (4)).

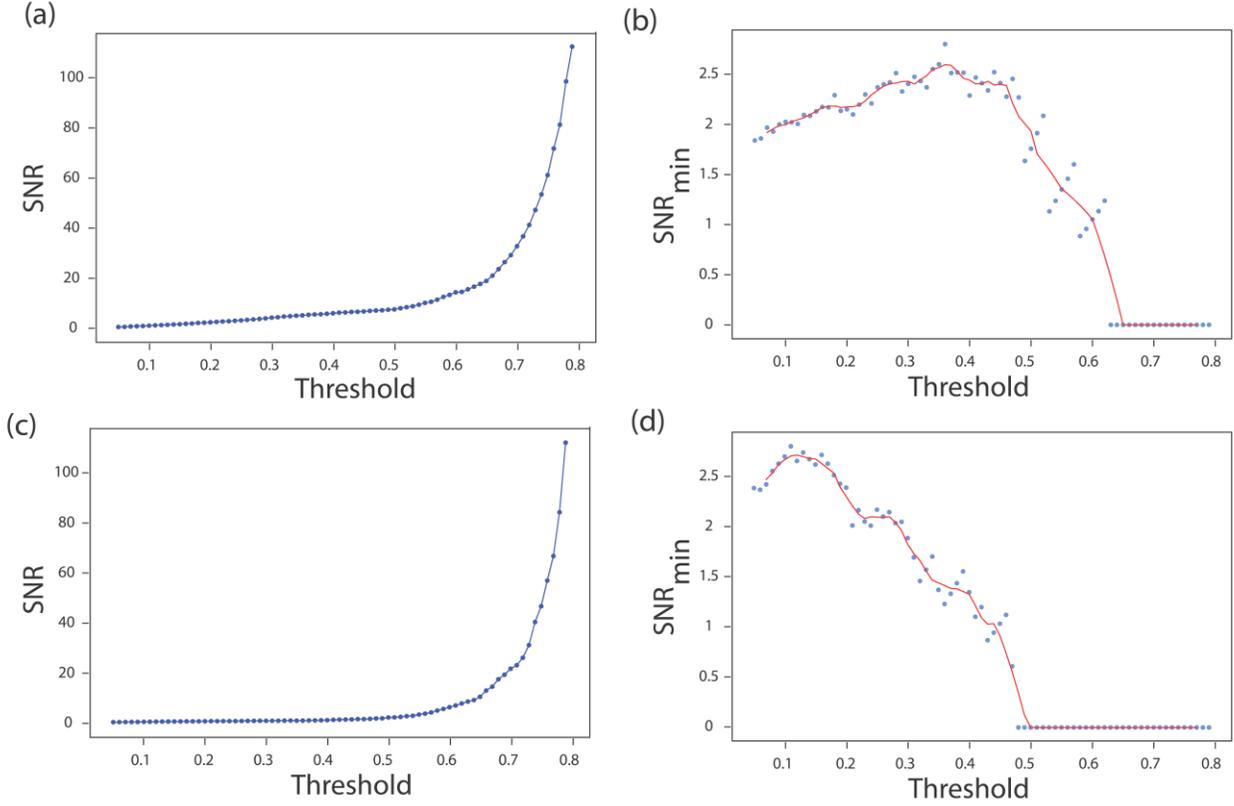

**Fig. 5.** Threshold selection. **(a)** SNR as a function of the threshold for transformer encoder 7. **(b)** The minimal SNR, $SNR_{min}$, based on the label with the minimal number of appearances in matrices' diagonals, as a function of threshold for transformer encoder 7. **(c)** Similar to panel a for transformer encoder 6. **(d)** Similar to panel b for transformer encoder 6.

## 3. Applied Nodal Diagonal Connection (ANDC) pruning

Previous work on CNNs demonstrated how quantifying single filter performance enables pruning the FC layer connected to the output layer without impacting overall accuracy[12,

13]. Similarly, one can prune the FC layer connecting the seventh transformer encoder to the 100 output units (Fig. 2) without affecting the overall accuracy. This is achieved using the ANDC pruning technique which allows for the removal of at least $80\%$ of the weights while maintaining an accuracy of $\sim 0.809$. In ANDC pruning, each of the $256$ nodes is connected only to the output units corresponding to the diagonal elements of its SNP matrix. The advantage of ANDC over random pruning is evident in the following example. Using a threshold of $0.4$ results in an average of $5.6$ diagonal elements in the 256 SNP matrices, leading to approximately $94.4\%$ pruning, and achieved $\sim 0.77$ accuracy after a short retraining session of the pruned layer. In contrast, simulations with a similar percentage of random pruning, followed by retraining, result in a significantly lower accuracy of $\sim 0.53$. This discrepancy highlights the importance of pruning based on SNP characteristics.

Because ViT architectures primarily comprise repeated FC layers (in both FF and MHA sub-blocks), applying ANDC pruning to these layers can significantly reduce the overall number of weights in the network. To achieve this, we connect a pair of input and output nodes only if their diagonal SNP matrices share at least one common element (Fig. 6). This ANDC pruning strategy is applied to the Linear1 and Linear2 layers of the FF sub-block, as well as to the MHA sub-block. For the MHA sub-block, a classifier head is introduced at the output of the LayerNorm operation ($256$ output units), the Query, Key, and Value (QKV) layer ($256$ units for each matrix, totaling $768$ outputs before self-attention), and the projection layer ($256$ output units) (Fig. 1). After sequentially training each of these classifier heads, the diagonal elements of each SNP matrix is calculated for a given threshold and ANDC pruning is applied.

Given the residual connections in the CCT-7/3x1 architecture (Fig. 1), the question arises whether to include the skip connection input fields when calculating the SNP matrices. Analysis of the seventh encoder's projection layer reveals that the magnitude of the skip connection field is $\sim 8$ times larger than the projection layer output fields, despite representing the same signal propagated through the MHA sub-block. Including the skip connection input in the classifier head results in different clusters and statistical features. This suggests that the SNP clusters are primarily determined by the skip connections. Therefore, pruning the projection layer weights ($256 \times 256$) based on these clusters enhances the overall signal, ensuring coherence between the skip connection and the MHA output. Consequently, pruning the projection layer according to the skip connection clusters allows for a higher dilution rate while maintaining the original accuracy of 0.809.

Results demonstrate that after a short retraining period (several epochs) an architecture with an average of 80% pruning in the two FC layers of the MHA and FF subblocks of the seventh transformer block, recovers the original accuracy of ~0.809 (Table 3). This pruning reduces the number of parameters in these four FC layers from $256 \cdot 256 + 2 \cdot 256 \cdot 512 + 256 \cdot 768 \sim 524{,}000$ to ~93,000.

Extending this ANDC pruning method to other transformer encoders while preserving accuracy is straightforward and has been confirmed in limited cases, such as pruning the Linear1 layer in multiple encoders. However, determining the optimal pruning strategy for the entire architecture is a more complex task and requires further investigation.

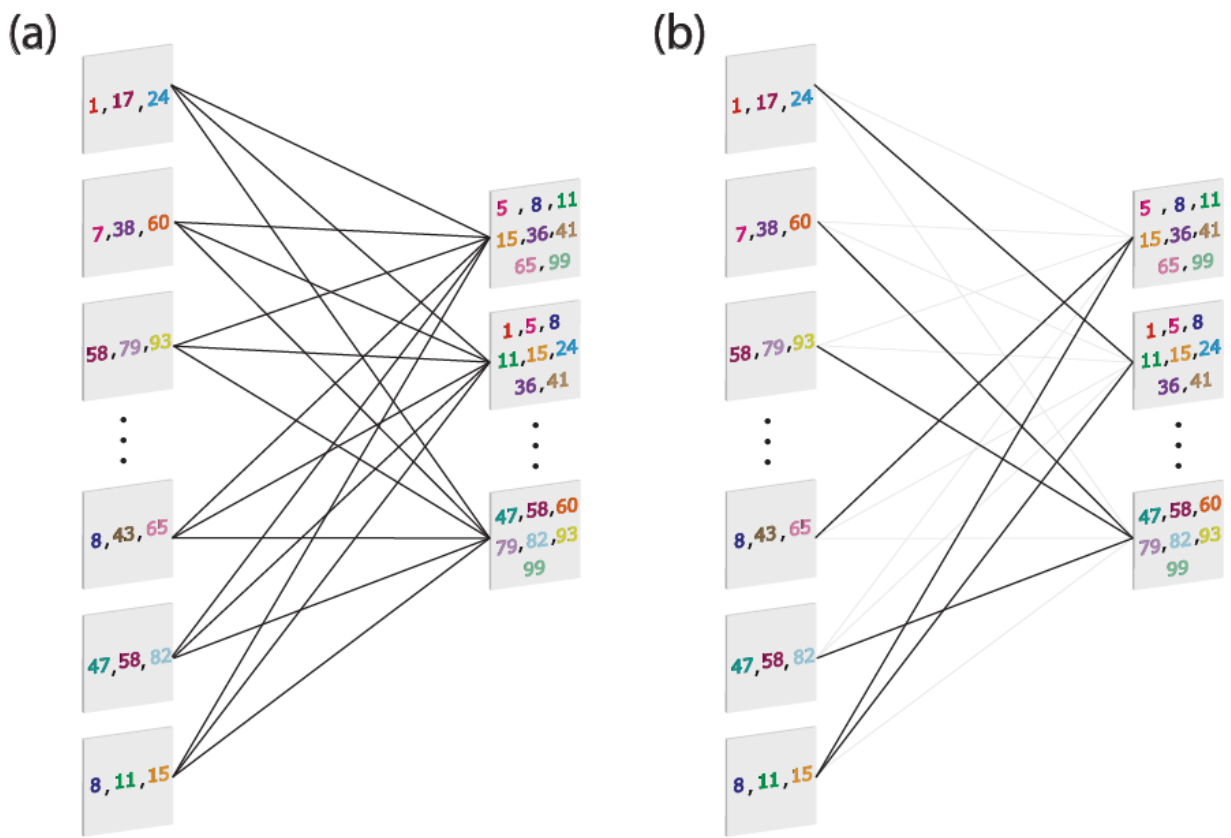

**Fig. 6:** ANDC pruning of a FC layer. **(a)** Scheme of SNP matrices (gray squares) of a FC layer of CCT-7/3x1, input (left) and output (right), with their above-threshold diagonal elements (numbers are color coded within the gray squares). **(b)** The same scheme after a significant ANDC pruning (pruned weights denoted by light gray lines), where nodes are connected only if their SNP matrices share at least one diagonal element.

| Layer | $Th$ | $Acc$ | $N_c$ | $C_s$ | $diag$ | $n$ | Dilution |
|---|---|---|---|---|---|---|---|
| CCT-7/3x1 - Block 7 ||||||||
| QKV MHA 256 input | 0.35 | 0.796 | 2.82 | 1.57 | 4.43 | 182.5 | 0.85 |
| QKV MHA 768 output | 0.35 | 0.797 | 2.28 | 1.63 | 3.72 | 201.6 | |
| Linear MHA 256 input | 0.3 | 0.783 | 2.71 | 1.52 | 4.12 | 137.3 | 0.77 |
| Linear MHA 256 output | 0.3 | 0.804 | 4.56 | 2.30 | 10.49 | 366.7 | |
| Linear1 256 input | 0.35 | 0.805 | 3.53 | 1.96 | 6.92 | 234.4 | 0.83 |
| Linear1 512 output | 0.35 | 0.797 | 1.48 | 1.49 | 2.21 | 79.6 | |
| Linear2 512 input | 0.35 | 0.797 | 1.48 | 1.49 | 2.21 | 79.6 | 0.80 |
| Linear2 256 output | 0.35 | 0.811 | 4.31 | 2.24 | 9.65 | 258.4 | |

**Table 3.** ANDC pruning of two FC layers of the MHA and the FF sub-blocks of the 7[th] block transformer, without affecting the accuracy, 0.809, of the CCT-7/3x1. The statistical features (labels as in Table 1) of the SNP matrices of the four input and output layers, for the four pruned FC layers, given their thresholds, $Th$, and resulted dilution (numbers are rounded).

## 4. Modus Vivendi among Multi-Head Attention

To investigate the interplay among the attention heads in the MHA sub-block, a trained CCT-7/3x1 architecture is analyzed. The weights of the first $m$ trained transformer encoders and the MHA layers of the $m + 1$ transformer encoder remained unchanged (frozen). The output of the scaled dot-product attention of the $m + 1$ MHA, $256 \times 256$ (Fig. 1), was then connected to a randomly initialized classifier head, comprising sequence pooling and an FC layer. This classifier head is trained to minimize the loss, similar to the procedure illustrated in Fig. 2 (middle). For the seventh MHA ($m = 6$), this procedure results in an accuracy of $0.786$. Next, the $256$ SNP matrices are calculated using the trained classifier head, silencing all weights in the FC layer except those ($100$) connected to a specific node (similar to Fig. 2, right). Next, the training inputs are presented to the network, processed by the first $m$ transformer encoders, the $m + 1$ MHA (excluding the projection layer), and the trained classifier head. This isolates the influence of a single node on the $100$ output units, allowing for the analysis of clusters and $noise_E$ within the resulting $100 \times 100$ matrix (similar to Fig. 2, right and Fig. 3a). The statistical properties of each of the four attention heads are inferred from their respective $64$ SNP matrices among the $256$, $[(h-1) \cdot 64 + 1, \ h \cdot 64]$ for $h = 1, 2, 3$ and $4$, focusing on the cumulative number of appearances of each label in the clusters (Fig. 7).

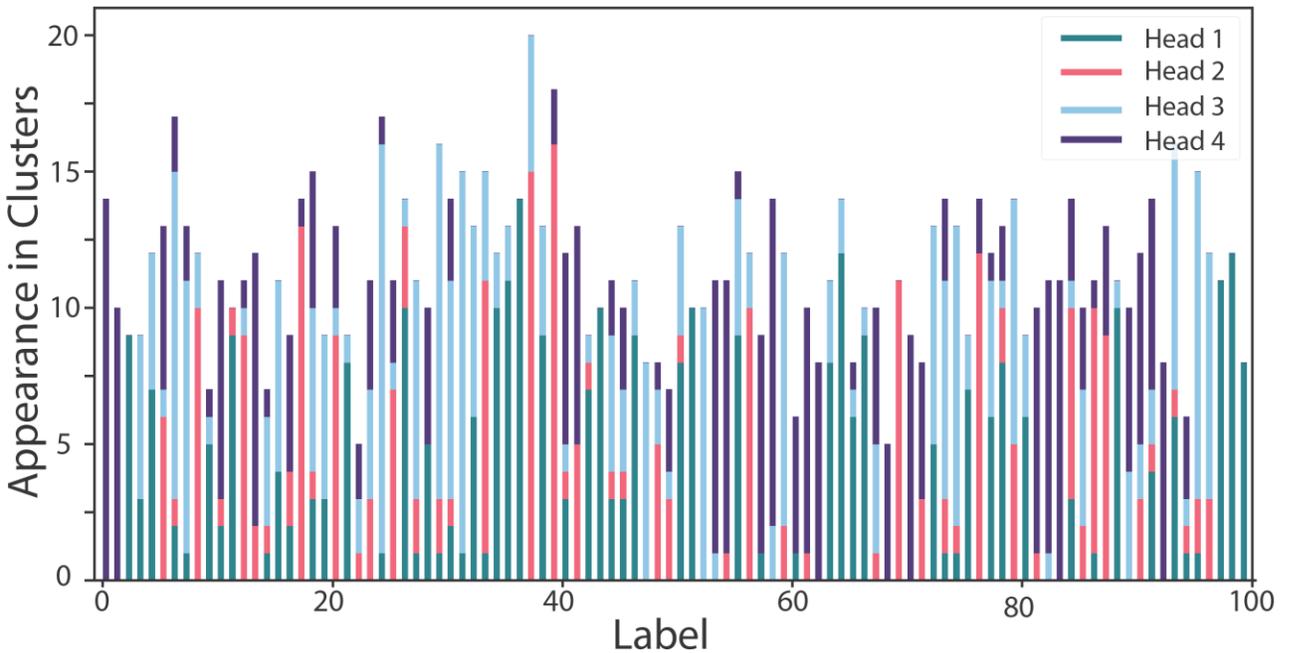

**Fig. 7.** Labels' cluster appearances for each one of the $4$ heads, measured by a trained classifier head connected to at the output of the scaled dot-product attention of the MHA of the $7^{th}$ transformer encoder. For each label, the accumulated number of its appearances as a diagonal cluster element in the $64$ SNP matrices belonging to a head is presented (color coded).

Each head is then assigned the labels whose cumulative cluster appearances in the diagonal elements of its $64$ SNP matrices are at least twice that of any other of the three heads (Fig. 8). Among the $100$ labels, $74$ meet this criterion which are split to $23, 15, 15$ and $21$ labels for heads $1, 2, 3$ and $4$, respectively (Table 4). This criterion effectively identifies the subset of labels recognized by each head. Remarkably, the test accuracies for these labels within their respective heads are consistently higher than those observed in the other three heads (Fig. 8). While the correspondence between diagonal appearances and accuracies is strong, it is not perfect. For instance, some labels exhibit relatively high test accuracies in a given head despite having minimal appearances in its diagonal SNP matrices (Fig. 8). This rare phenomenon can be attributed to the influence of noise ($n$), outside the clusters, and the omission of sub-threshold matrix elements. Furthermore, the normalization of SNP matrices by their maximum values neglects their relative amplitudes, leading to non-uniform contributions to each head's accuracy.

These results reveal a spontaneous symmetry breaking in label recognition among the four heads. Each head specializes in a subset of labels through cooperation among its SNPs, effectively becoming an expert in recognizing its labels. This spontaneous symmetry breaking develops progressively along the transformer blocks. For example, in the MHA of the sixth transformer encoder, only $39$ labels are distinctly recognized by the four heads (Table 4). Three heads are in their initial specialization phase, recognizing only $4-5$ labels each, whereas the fourth head has already become an expert on $23$ labels. In earlier encoder blocks, all four heads are in this initial professionalization learning phase.

This spontaneous symmetry breaking is also evident in the distribution of labels across the $20$ super-classes of CIFAR-100[10], each containing five labels. Each head in the seventh transformer encoder specializes in labels belonging to a few super-classes, demonstrating a tendency far beyond random label selection. Simulations with random label selection reveal significantly lower probabilities for observing such super-class occupation compared to the actual head specialization (Table 4). For instance, the first head specializes in $23$ labels (Fig. 8 and Table 4), fully occupying two super-classes. In contrast, the probability of this occurring with random label selection is negligible (Table

4). This tendency to occupy super-classes diminishes in three of the heads in the sixth transformer encoder, consistent with their early specialization stage (Table 4).

The modus vivendi among the MHA heads enhances the accuracy of labels with spontaneous symmetry breaking (Fig. 9). For a pre-trained CCT-737x1 architecture on CIFAR-100, the accuracy per label was measured. The averaged accuracy for labels with spontaneous symmetry breaking of $0.82$ is superior to $0.78$ for the rest (Fig. 9, upper panel). Similarly, an average accuracy for the subset labels with symmetry breaking in the 6$^{th}$ transformer block of $0.836$ appears in comparison to $0.79$ for the other labels (Fig. 9, lower panel). The average accuracy for labels with spontaneous symmetry breaking is further enhanced to ~$0.857$ for labels with symmetry breaking both in the 6$^{th}$ and the 7$^{th}$ transformer blocks.

| Block 7 Head (total) | Appearance No. Labels/Coarse Label | | | | |
|---|---|---|---|---|---|
| | 1/5 | 2/5 | 3/5 | 4/5 | 5/5 |
| 1 (23) | 3 | 3 | 0 | 1 | 2 |
| 2 (15) | 3 | 1 | 2 | 1 | 0 |
| 3 (15) | 3 | 2 | 0 | 2 | 0 |
| 4 (21) | 6 | 0 | 2 | 1 | 1 |

| 50,000 Random Samples No. labels | Av. Appearance No. Labels/Coarse Label | | | | |
|---|---|---|---|---|---|
| | 1/5 | 2/5 | 3/5 | 4/5 | 5/5 |
| 25 | 8.1 | 5.4 | 1.7 | 0.25 | 0.015 |
| 20 | 8.4 | 4.2 | 0.95 | 0.10 | 0.004 |
| 15 | 8.1 | 2.8 | 0.44 | 0.03 | 0.0007 |
| 10 | 6.8 | 1.4 | 0.13 | 0.005 | 0 |
| 5 | 4.2 | 0.37 | 0.013 | 0 | 0 |

| Block 6 Head (total) | Appearance No. Labels/Coarse Label | | | | |
|---|---|---|---|---|---|
| | 1/5 | 2/5 | 3/5 | 4/5 | 5/5 |
| 1 (4) | 4 | 0 | 0 | 0 | 0 |
| 2 (26) | 3 | 4 | 2 | 1 | 1 |
| 3 (5) | 5 | 0 | 0 | 0 | 0 |
| 4 (4) | 2 | 1 | 0 | 0 | 0 |

**Table 4.** Super-classes recognized by the four heads. The CIFAR-100 dataset is comprised of 100 labels where every 5 labels constitute a specific super-class. Left table presents for each head of the 7$^{th}$ MHA, the number of occupied super-classes and their filling, $\frac{i}{5}$ ($i = 1, 2, 3, 4$ and $5$). Middle table presents the number of occupied super-classes and their filling obtained in simulations for a given number of random selected labels,

averaged over 50,000 samples. Right table presents the same as the top table, but for the 6$^{th}$ MHA.

This spontaneous symmetry breaking is also confirmed on the Flowers-102 dataset[11], which comprises 102 labels and images of size $224 \times 224$ (Table 5). This dataset was trained on the CCT-7/7x2 architecture, which begins with two CLs of sizes 7 and 3, followed by seven transformer encoders identical to those in CCT-7/3x1[9] (Fig. 1). This model achieves an accuracy of ~0.972 after 300 epochs. Similar to CIFAR-100, spontaneous symmetry breaking occurs among the four heads in label recognition (Fig. 8), but it emerges in an earlier MHA block (Table 5). In the sixth MHA, three heads specialize in recognizing significant subsets of labels, with this specialization increasing in the seventh MHA through the expansion of subset sizes and the inclusion of the fourth head.

The reported symmetry breaking of a label (Fig. 8 and Table 4) relies on a criterion that the cumulative cluster appearances in the diagonal elements of the 64 SNP matrices of a head is at least greater than twice that of any other of the three heads. The sensitivity of presented results for this predefined criterion is examined where a label with symmetry breaking is defined if its cumulative appearances is greater than $Th_{Ratio}$ (instead of $Th_{Ratio} = 2$) in comparison to the other three heads (Fig. 10). Results indicate that significant symmetry breaking emerges only at the 7$^{th}$ transformer block, where ~70 labels with symmetry breaking appears for $Th_{Ratio}$ in the range $(1.5, 2.5)$, where for lower transformer blocks this criterion is fulfilled for $Th_{Ratio} < 1.4$. In addition, the limit $Th_{Ratio} \to 1$ typically identifies fluctuations among the heads with low accuracies per label (Table 1), rather than a signature for a symmetry breaking. Similarly, for blocks 4 and 5, the lower values of labels with symmetry breaking for $Th_{Ratio} > 2$ is a result of fluctuations (in the order of $\sqrt{No. of\ labels}$) and cannot be a signature for a spontaneous symmetry breaking.

For images recognized by a specific head, the average scaled dot-product attention output fields (attention map) exhibit high similarity to the input image (Fig. 11, first row), in comparison to the other three heads (Fig. 11, second row). This similarity is further enhanced with ANDC pruning (Fig. 11, first row, right image). For another example, the average scaled dot-product attention output fields of a head recognizing a specific image

(Fig. 11, third row, second from right) dominate the output field across all head dimensions (Fig. 11, third row, second from left). However, these qualitative similarities are infrequent, observed only in a relatively small fraction of the examined input images, and thus cannot serve as a quantitative measure for the underlying MHA learning mechanism.

| Flowers-102 CCT-7/7x2 | | | | | | |
|---|---|---|---|---|---|---|
| MHA | Acc | Head 1 | Head 2 | Head 3 | Head 4 | Total |
| 7 | 0.97 | 18 | 20 | 16 | 14 | 68 |
| 6 | 0.93 | 17 | 13 | 7 | 13 | 50 |
| 5 | 0.88 | 4 | 8 | 7 | 5 | 24 |
| 4 | 0.81 | 0 | 3 | 0 | 4 | 7 |

**Table 5.** The emergence of symmetry breaking in the recognition of subsets of Flower-102 labels by the four heads of CCT-7/7x2 architecture along the $[4, 7]$ MHAs and their increased accuracy, $Acc$.

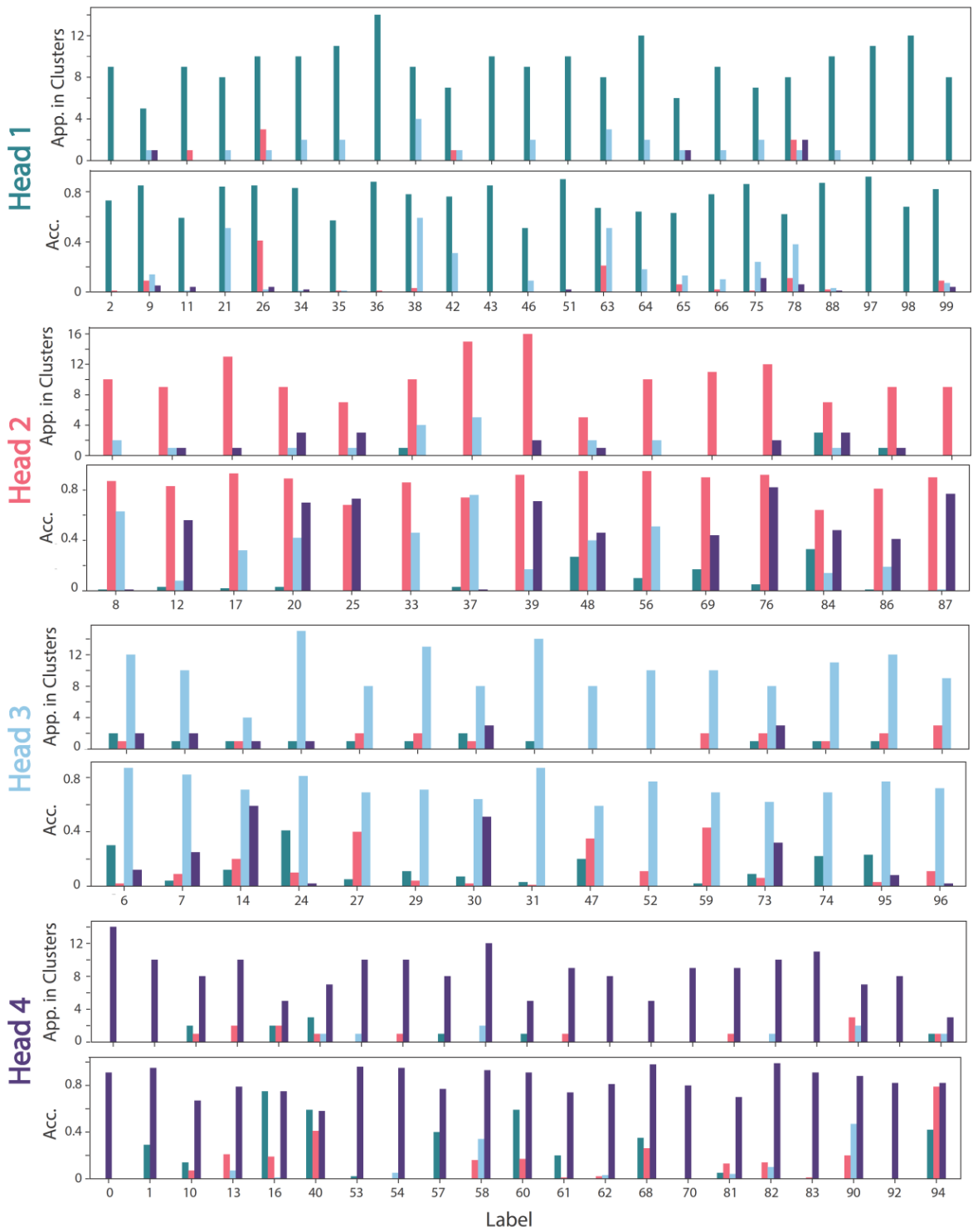

**Fig. 8.** Spontaneous symmetry breaking among the labels recognized by each one of the four heads of the 7th MHA. Each one of the four heads (color coded as in Fig. 7) denotes the labels that their accumulative appearance (app.) in clusters in its $64$ SNP matrices are at least twice more than for each one of the rest three heads (upper graph for each head). For each head and its corresponding labels, the lower graph presents the test accuracy for each one of the four heads, color coded, respectively.

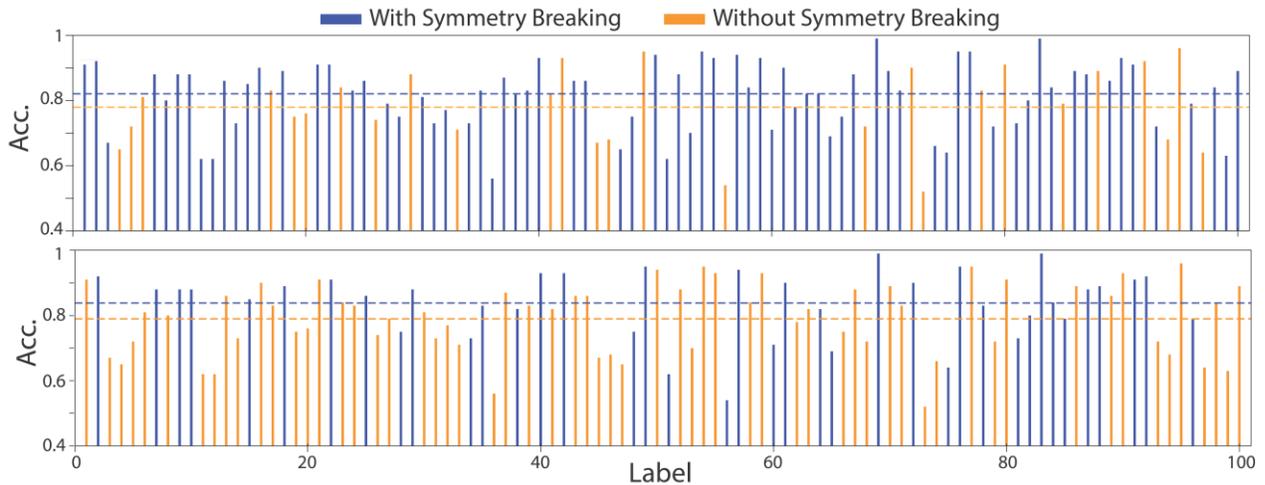

**Fig. 9.** The accuracy per label, $ACC$, is measured on the output layer of a pre-trained CCT-7/3x1 architecture on the CIFAR-100 dataset. Upper panel: Labels with symmetry breaking (blue) and labels without symmetry breaking (orange) (similar to Table 4) for the 7th MHA and their average values (horizontal dashed-line). Lower panel: The same $ACC$ per label as in the upper panel, where labels with symmetry breaking in the 6th MHA (Table 4) (orange) and the rest (blue) and their average values (horizontal dashed-line).

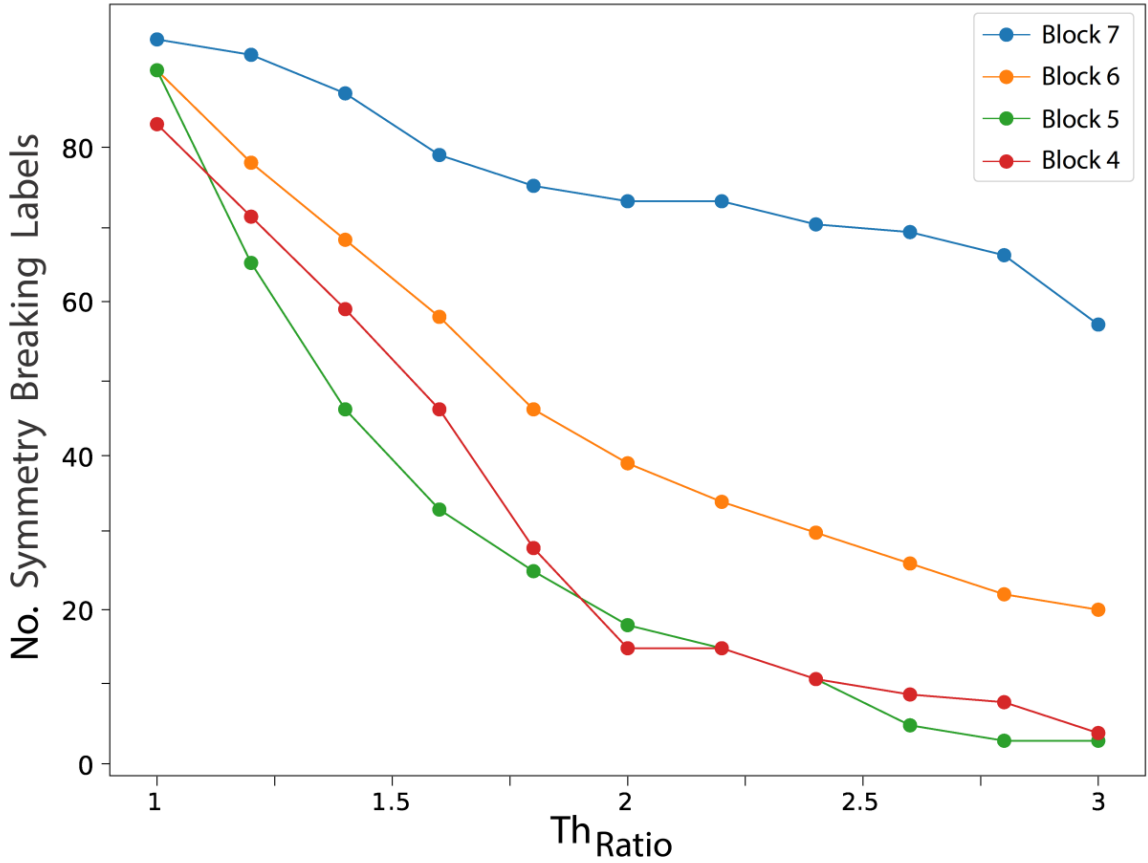

**Fig. 10.** The total number of labels, in all the four heads, with spontaneous symmetry breaking as a function of $Th_{Ratio}$, for blocks 4, 5, 6 and 7 (color coded) for CCT-7/7x1 trained on CIFAR-100 (similar to Fig. 8). For a given head in a transformer block, a label with spontaneous symmetry breaking is defined if the cumulative cluster appearances in the diagonal elements of its 64 SNP matrices are greater by a factor of $Th_{Ratio}$ than of any of the other three heads.

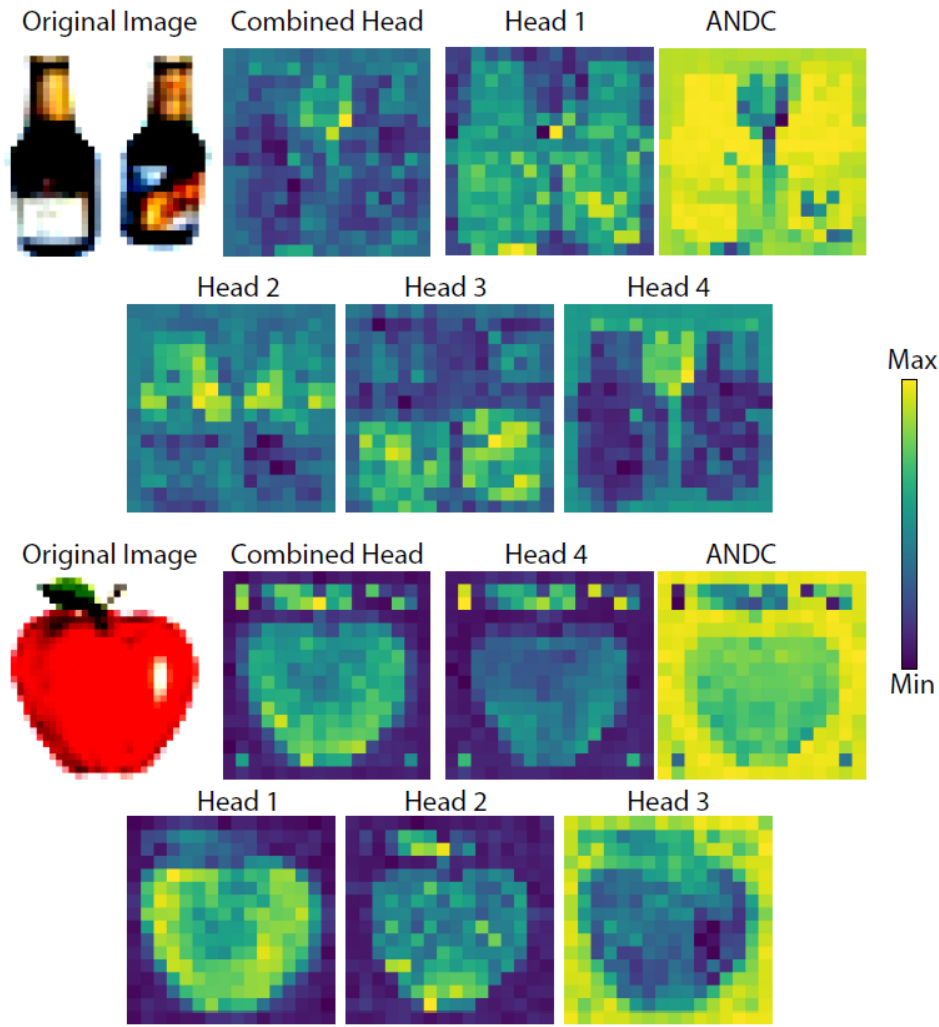

**Fig. 11.** Scaled dot-product attention output fields for two specific test images which correctly predicted. Top row: the image (left), the average attention output field across all heads' dimensions (second from left), the attention output fields of a head who recognizes the label (second from the right), the average attention output field across all heads with ANDC dilution (right). Second row: Attention output fields across each of the rest three heads.

## 5. Discussion

CNNs and ViT architectures were developed and optimized to solve complex vision classification tasks from distinct perspectives. CNNs leverage CLs that progressively expand their receptive field, starting from local correlations and gradually incorporating broader context to identify the correct label. In contrast, ViT architectures, based on repeated transformer encoders, capture long-range correlations across the entire image primarily through FC layers.

Given the fundamental differences in their design, it is assumed that CNNs and ViT architectures rely on distinct learning mechanisms. However, contrary to this intuition and common belief, this study demonstrated that both types of architectures share a unified underlying learning mechanism. This finding stems from our proposed method for quantifying the SNP of each node within ViT architectures, encompassing both FF and MHA subblocks, analogous to single filter performance analysis in CNNs. Results indicated that both filters in CNNs and nodes in ViT identify small clusters of possible output labels, with additional noise represented as labels outside these clusters (Fig. 3). These features become increasingly refined along the network, enhancing SNR and leading to improved accuracy along layers in CNNs and transformer encoder blocks in ViT architectures (Table 1).

Furthermore, based on SNP properties, a modus vivendi was observed among the heads of the MHA mechanism, characterized by spontaneous symmetry breaking. Each head places its attention on a different subset of labels, becoming an expert in their recognition, achieving significantly higher accuracy compared to other heads. This spontaneous symmetry breaking arises from cooperation among the nodes within each head, strengthens along transformer blocks, and constitutes the underlying learning and classification mechanism of MHA. This quantitative modus vivendi among the MHA improves the infrequent overlap between the average scaled dot-product attention output fields and a correctly classified image or its selected part (Fig. 11).

For the CIFAR-100 dataset comprising 100 labels, organized into 20 super-classes (each with five labels), the symmetry breaking among the MHA extends beyond individual labels to encompass super-classes. Each attention head specializes in labels belonging to a limited number of super-classes, demonstrating a tendency far beyond random label selection (Table 4).

The MHA modus vivendi mechanism raises several research questions and potential practical implications. First, it suggests investigating the optimal number of attention heads as a function of the number of labels. For instance, the optimal number of heads for CIFAR-100 likely exceeds that for CIFAR-10, where each head would specialize in a very small number of labels. Indeed, training performance on CIFAR-10 on CCT-7/7x3 revealed that no single head dominates the accuracy of any label, as each head can specialize in many more than three labels, suggesting the potential for using a simpler MHA. Second, it may be beneficial to increase the number of heads along the transformer encoder blocks, as all heads are in the pre-specialization stage in earlier blocks (Tables 4 and 5). Third, the discovery of the underlying spontaneous symmetry breaking mechanism in MHA could lead to advanced backpropagation variants that explicitly enforce this symmetry breaking. For example, one could pre-assign subsets of labels (or super-classes) to each head in each encoder, with enhanced learning parameters for each head's subset. These subsets could be partially overlapping, randomly selected, or tailored based on label similarity. Preliminary results indicate that enhancing learning parameters for specific input label subsets associated with each head indeed enforces symmetry breaking, and might even increase accuracy (similar to Fig. 9). Finally, SNP provides an estimate of the average functionality of each node. Silencing or retraining specific nodes could enhance their cooperation and improve SNR. Notably, a small number of SNP matrices exhibit vanishing diagonals but significant noise, and their pruning does not affect overall accuracy. Therefore, pruning SNP matrices with relatively small maximum fields or retraining them could potentially enhance accuracy.

The demonstrated unified learning mechanism for CNNs and ViT architectures hints at a universal learning mechanism applicable to other architectures and applications beyond computer vision, such as natural language processing (NLP)[1]. Furthermore, this understanding leads to a quantitative pruning technique that significantly reduces the number of required network parameters without compromising accuracy. This reduction is expected to significantly decrease the computational cost of required floating-point operation in the feedforward and backpropagation steps, minimizing memory and energy consumption during training, testing, and deployment, particularly for resource-constrained end users. However, optimal utilization of pruned architectures might necessitate specialized hardware with asynchronous dynamics, where each node is individually controlled, inspired by biological neural networks[14-17]. It is important to note that the proposed universal unified learning mechanism is robust to the choice of the optimizer, showing consistent results with both stochastic gradient descent (SGD) with Nesterov momentum[18] and AdamW[19].

The ANDC pruning of CCT-7/3x1 was primarily demonstrated for the seventh transformer encoder and warrants further investigation in the other six encoders and across different ViT architectures[20-23]. The ~80% pruning of the seventh CCT-7/3x1 block without impacting accuracy is modest compared to the reported pruning of VGG-16, which can exceed 99% in some FC layers. However, VGG-16, with $\sim 16\,M$ parameters, achieves only ~0.75 accuracy on CIFAR-100, whereas CCT-7/3x1 attains 0.809 accuracy with only $\sim 3.7\,M$ parameters. Significantly greater ANDC pruning potential is anticipated without compromising accuracy when classifying larger images and datasets using ViT architectures or NLP transformers[24-27] with billions of parameters.

An intriguing approach for implementing ANDC pruning is to calculate the SNP while excluding skip connections from the previous transformer sub-block (Fig. 1). This typically results in considerably smaller diagonals and reduced noise, potentially enhancing ANDC pruning. For example, in the seventh transformer encoder, the average ~15 number of diagonals and ~417 noise (Table 1) decrease to ~8 and ~326, respectively when excluding skip connections. This effect might arise from the biases introduced by skip connections within the inter-block fields, leading to larger clusters, diagonals, and noise. Further research is required to explore the efficiency of this ANDC pruning variant.

The ANDC pruning technique also enables the evaluation of learning architectures using a new benchmark focused on minimizing the number of trained parameters for a given dataset and accuracy. This contrasts with traditional benchmarks that aim to maximize accuracy for a given dataset using advanced architectures. For instance, the goal could be to find the architecture with the fewest trained parameters that achieves 0.9 accuracy on CIFAR-100. This type of benchmark could reveal quantitative relationships between dataset size, number of labels, and the minimum number of required parameters, potentially establishing bounds for optimal pruning and lower limits for memory and energy consumption in classification tasks with specific accuracy requirements. While such goals have been analytically achieved for learning random input/output examples using simple architectures, where the capacity per weight is 2 for the perceptron and $\log_2(K)$ for multi-layer perceptrons with one hidden layer and $K$ hidden units[28-30], similar analytical results for deep learning with non-random data seem currently unattainable. However, their numerical estimation remains a plausible avenue for exploration.

A potential drawback of ANDC pruning is the need for initial training to maximize accuracy, followed by retraining to maintain accuracy after pruning. This double training procedure might be acceptable for providers of ViT software or hardware, as end users benefit from the highly pruned architecture, which minimizes memory and energy consumption. However, an artificial version of ANDC pruning, A-ANDC, could circumvent this double training requirement. A-ANDC pre-assigns a diagonal set of labels to each node, ensuring that each label appears almost evenly in the nodal matrices of each layer (Fig. 4a). The artificially selected diagonal sizes can be based on prior knowledge or adaptively increased until the desired accuracy is achieved. This artificial pruning approach has proven effective for CNNs and merits further investigation in ViT architectures.

The presented underlying mechanism for ViT architectures also offers guidance on the interplay between the number of SNP matrices ($N_M$) in MHA, their average diagonal size ($D_A$), and the number of labels ($N_L$). Each label should appear multiple times in the $N_M$ SNP matrices to constitute a signal (Eq. (1)), $N_M \cdot D_L \gg N_l$. Therefore, for instance, for $N_l = 10{,}000$, and $N_M = 256$ as in CCT-7/7x1, the diagonals would need to be considerably larger than 40 and contain significant clusters and noise, $n$, which is improbable. Consequently, a wider MHA with a larger $N_M$ is recommended.

Finally, the emergence of symmetry breaking along transformer blocks in ViT architectures is reminiscent of symmetry breaking phenomena in physics, particularly in statistical physics during phase transitions[31, 32]. Progressing through the transformer blocks can be viewed as analogous to decreasing the temperature toward a critical point in a second-order phase transition, where the correlation length diverges is replaced by proceeding along the transformer blocks. However, it remains unclear whether the spontaneous symmetry breaking in transformer blocks exhibits characteristics of a second-order phase transition, with increasing correlation length among MHA units along earlier blocks. Owing to the constraints of the current study, which investigated relatively shallow ViT architectures (seven transformer blocks) and small datasets, it is difficult to answer these questions. Therefore, further investigation with larger input images, larger MHAs, and deeper architectures is needed to elucidate the nature of this symmetry breaking phenomenon.

**Acknowledgements**

The work is supported by the Israel Science Foundation [grant number 346/22].

**Appendix**

*Dataset and preprocessing:* The datasets used in this study are CIFAR-100[10] and Flowers-102[11]. Each pixel value was normalized by subtracting the mean and dividing by the standard deviation of its image.

*Optimization:* The Label Smoothing Cross-Entropy function was selected for the classification task and minimized using the stochastic gradient descent algorithm[5, 33]. The maximal accuracy was determined by searching through the hyper-parameters (see below). The L2 regularization method[34] was applied. The dataset CIFAR-100 was tested on the CCT-7/3x1 architecture[9], consists of a CL of size 3 and 7 transformer encoders (Fig. 1), where Flowers-102 dataset was tested on CCT-7/7x2 architecture[9], consists of two CLs of sizes 7 and 3 and 7 transformer encoders (Fig. 1).

*Hyper-parameters:* The hyper-parameters $\eta$ (learning rate) and $\alpha$ (L2 regularization) were optimized for offline learning, using a mini-batch size of 100 inputs. The learning-rate decay schedule was also optimized. A linear scheduler was applied such that it was multiplied by the decay factor, $q$, every $\Delta t$ epochs, and is denoted below as $(q, \Delta t)$. Statistics for each data point were computed based on at least three trials, with fluctuations being less than 0.5%.

The hyper parameters used for Tables 1 and 2 are $\eta = 3e-4$ and $\alpha = 4e-2$, using the decay schedule $(q, \Delta t) = (0.8, 10)$ and The architecture was trained for 100 epochs.

The hyper parameters used for the training of the classifier head in Table 3 are $\eta = 1e-4$ and $\alpha = 5e-2$, using the decay schedule $(q, \Delta t) = (0.78, 10)$. The architecture was trained for 30 epochs. The pruned architecture was trained for a few epochs using $\eta = 5e-4$ and $\alpha = 1e-2$ for the last transformer encoder and $\eta = 1e-6$ and $\alpha = 1e-2$ for the rest trained architecture.

The hyper parameters used for the training the classifier head in Table 5 are $\eta = 1e-3$ and $\alpha = 5e-2$, using the decay schedule $(q, \Delta t) = (0.78, 10)$, and the architecture was trained for 30 epochs.

The hyper parameters used for the training of the classifier head in Fig. 3 ((a) and (b)), Fig. 4 and Fig. 5 are $\eta = 5e-3$ and $\alpha = 5e-2$, using the decay schedule $(q, \Delta t) = (0.78, 10)$, and the architecture was trained for 50 epochs.

The hyper parameters used for the training of the classifier head in Table 4, Fig. 7 and Fig. 8 are $\eta = 1e-3$ and $\alpha = 5e-2$, using the decay schedule $(q, \Delta t) = (0.78, 10)$, and the architecture was trained for 30 epochs.

*Statistics:* Statistics for all results were obtained using at least three samples and the standard division was less than $0.5\%$ for all the results.

*Hardware and software*: We used Google Colab Pro and its available GPUs. We used Pytorch for all the programming processes.

[1] A. Vaswani, Attention is all you need, Advances in Neural Information Processing Systems, (2017).
[2] A. Dosovitskiy, An image is worth 16x16 words: Transformers for image recognition at scale, arXiv preprint arXiv:2010.11929, (2020).
[3] M. Tan, Q. Le, Efficientnet: Rethinking model scaling for convolutional neural networks, in: International conference on machine learning, PMLR, 2019, pp. 6105-6114.
[4] K. Simonyan, A. Zisserman, Very deep convolutional networks for large-scale image recognition, arXiv preprint arXiv:1409.1556, (2014).
[5] K. He, X. Zhang, S. Ren, J. Sun, Deep residual learning for image recognition, in: Proceedings of the IEEE conference on computer vision and pattern recognition, 2016, pp. 770-778.
[6] M. Raghu, T. Unterthiner, S. Kornblith, C. Zhang, A. Dosovitskiy, Do vision transformers see like convolutional neural networks?, Advances in neural information processing systems, 34 (2021) 12116-12128.
[7] J.-B. Cordonnier, A. Loukas, M. Jaggi, On the relationship between self-attention and convolutional layers, arXiv preprint arXiv:1911.03584, (2019).
[8] M. Chen, A. Radford, R. Child, J. Wu, H. Jun, D. Luan, I. Sutskever, Generative pretraining from pixels, in: International conference on machine learning, PMLR, 2020, pp. 1691-1703.
[9] A. Hassani, S. Walton, N. Shah, A. Abuduweili, J. Li, H. Shi, Escaping the big data paradigm with compact transformers, arXiv preprint arXiv:2104.05704, (2021).
[10] A. Krizhevsky, G. Hinton, Learning multiple layers of features from tiny images, (2009).
[11] M.-E. Nilsback, A. Zisserman, Automated flower classification over a large number of classes, in: 2008 Sixth Indian conference on computer vision, graphics & image processing, IEEE, 2008, pp. 722-729.
[12] Y. Meir, Y. Tzach, S. Hodassman, O. Tevet, I. Kanter, Towards a universal mechanism for successful deep learning, Scientific Reports, 14 (2024) 5881.
[13] Y. Tzach, Y. Meir, O. Tevet, R.D. Gross, S. Hodassman, R. Vardi, I. Kanter, The mechanism underlying successful deep learning, arXiv preprint arXiv:2305.18078, (2023).
[14] Y. Meir, I. Ben-Noam, Y. Tzach, S. Hodassman, I. Kanter, Learning on tree architectures outperforms a convolutional feedforward network, Sci Rep-Uk, 13 (2023) 962.
[15] S. Sardi, R. Vardi, Y. Meir, Y. Tugendhaft, S. Hodassman, A. Goldental, I. Kanter, Brain experiments imply adaptation mechanisms which outperform common AI learning algorithms, Sci Rep-Uk, 10 (2020) 6923.
[16] Y. Meir, O. Tevet, Y. Tzach, S. Hodassman, I. Kanter, Role of delay in brain dynamics, Physica A: Statistical Mechanics and its Applications, 654 (2024) 130166.
[17] O. Tevet, R.D. Gross, S. Hodassman, T. Rogachevsky, Y. Tzach, Y. Meir, I. Kanter, Efficient shallow learning mechanism as an alternative to deep learning, Physica A: Statistical Mechanics and its Applications, 635 (2024) 129513.
[18] I. Sutskever, J. Martens, G. Dahl, G. Hinton, On the importance of initialization and momentum in deep learning, in: International conference on machine learning, 2013, pp. 1139-1147.
[19] I. Loshchilov, Decoupled weight decay regularization, arXiv preprint arXiv:1711.05101, (2017).
[20] Z. Liu, Y. Lin, Y. Cao, H. Hu, Y. Wei, Z. Zhang, S. Lin, B. Guo, Swin transformer: Hierarchical vision transformer using shifted windows, in: Proceedings of the IEEE/CVF international conference on computer vision, 2021, pp. 10012-10022.
[21] N. Kitaev, Ł. Kaiser, A. Levskaya, Reformer: The efficient transformer, arXiv preprint arXiv:2001.04451, (2020).
[22] K. Han, A. Xiao, E. Wu, J. Guo, C. Xu, Y. Wang, Transformer in transformer, Advances in neural information processing systems, 34 (2021) 15908-15919.
[23] K. Han, Y. Wang, H. Chen, X. Chen, J. Guo, Z. Liu, Y. Tang, A. Xiao, C. Xu, Y. Xu, A survey on vision transformer, IEEE transactions on pattern analysis and machine intelligence, 45 (2022) 87-110.

[24] J.D.M.-W.C. Kenton, L.K. Toutanova, Bert: Pre-training of deep bidirectional transformers for language understanding, in: Proceedings of naacL-HLT, Minneapolis, Minnesota, 2019, pp. 2.
[25] Y. Liu, Roberta: A robustly optimized bert pretraining approach, arXiv preprint arXiv:1907.11692, 364 (2019).
[26] J. Achiam, S. Adler, S. Agarwal, L. Ahmad, I. Akkaya, F.L. Aleman, D. Almeida, J. Altenschmidt, S. Altman, S. Anadkat, Gpt-4 technical report, arXiv preprint arXiv:2303.08774, (2023).
[27] C. Zhang, C. Zhang, S. Zheng, Y. Qiao, C. Li, M. Zhang, S.K. Dam, C.M. Thwal, Y.L. Tun, L.L. Huy, A complete survey on generative ai (aigc): Is chatgpt from gpt-4 to gpt-5 all you need?, arXiv preprint arXiv:2303.11717, (2023).
[28] I. Kanter, Information theory of a multilayer neural network with discrete weights, Europhys Lett, 17 (1992) 181.
[29] E. Barkai, D. Hansel, I. Kanter, Statistical mechanics of a multilayered neural network, Phys Rev Lett, 65 (1990) 2312.
[30] G. Mitchison, R. Durbin, Bounds on the learning capacity of some multi-layer networks, Biol Cybern, 60 (1989) 345-365.
[31] A. Barra, G. Genovese, P. Sollich, D. Tantari, Phase transitions in restricted Boltzmann machines with generic priors, Phys Rev E, 96 (2017) 042156.
[32] E. Agliari, L. Albanese, A. Barra, G. Ottaviani, Replica symmetry breaking in neural networks: a few steps toward rigorous results, Journal of Physics A: Mathematical and Theoretical, 53 (2020) 415005.
[33] J. Schmidhuber, Deep learning in neural networks: An overview, Neural networks, 61 (2015) 85-117.
[34] C. Cortes, M. Mohri, A. Rostamizadeh, L2 regularization for learning kernels, arXiv preprint arXiv:1205.2653, (2012).